\documentclass[letterpaper, 10 pt, conference]{ieeeconf}  % Comment this line out if you need a4paper

\IEEEoverridecommandlockouts                              % This command is only needed if 
\usepackage{graphicx} 
\usepackage{tabularx}
\usepackage{xcolor}
\usepackage{amsfonts}
\usepackage{amsmath}

\usepackage{cite}

\title{%\LARGE \bf
TraveLLM: Could you plan my public transit alternatives in face of a network disruption?
}
\author{
Bowen Fang, %$^{*}$, 
Zixiao Yang, %$^{*}$, 
 Xuan Di \textit{Member, IEEE}% <-this % stops a space
\thanks{\textit{(Corresponding author: Bowen Fang)}}
\thanks{Bowen Fang and Zixiao Yang are with the Department of Industrial Engineering and Operations Research,
        Columbia University, New York City, NY, 10027 USA
        (e-mail: bf2504@columbia.edu, zy2531@columbia.edu). %\textit{(*: Bowen Fang and Zixiao Yang equally contributed to this work.)}
}%
\thanks{Xuan Di is with the Department of Civil Engineering and Engineering
Mechanics, Columbia University, New York, NY, 10027 USA, and also with
the Data Science Institute, Columbia University, New York, NY, 10027 USA
(e-mail: sharon.di@columbia.edu).}%
}

\begin{document}

\maketitle
\thispagestyle{empty}
\pagestyle{empty}

% TODO: TRR (transportation research record) public transit, alternative, disruption; transit alternatives 
%%%%%%%%%%%%%%%%%%%%%%%%%%%%%%%%%%%%%%%%%%%%%%%%%%%%%%%%%%%%%%%%%%%%%%%%%%%%%%%%
\begin{abstract}
Existing navigation systems often fail during urban disruptions, struggling to incorporate real-time events and complex user constraints, such as avoiding specific areas. We address this gap with TraveLLM, a system using Large Language Models (LLMs) for disruption-aware public transit routing. We leverage LLMs' reasoning capabilities to directly process multimodal user queries combining natural language requests (origin, destination, preferences, disruption info) with map data (e.g., subway, bus, bike-share). To evaluate this approach, we design challenging test scenarios reflecting real-world disruptions like weather events, emergencies, and dynamic service availability. We benchmark the performance of state-of-the-art LLMs, including GPT-4, Claude 3, and Gemini, on generating accurate travel plans. Our experiments demonstrate that LLMs, notably GPT-4, can effectively generate viable and context-aware navigation plans under these demanding conditions. These findings suggest a promising direction for using LLMs to build more flexible and intelligent navigation systems capable of handling dynamic disruptions and diverse user needs.

\end{abstract}

%%%%%%%%%%%%%%%%%%%%%%%%%%%%%%%%%%%%%%%%%%%%%%%%%%%%%%%%%%%%%%%%%%%%%%%%%%%%%%%%
\section{INTRODUCTION}

Urban mobility systems frequently encounter disruptions, ranging from severe weather events and infrastructure failures to public emergencies, posing persistent challenges to commuters \cite{touloumidis2025weather}. For instance, navigating New York City during a major storm involves contending with widespread transit shutdowns and localized hazards, demanding navigation solutions that extend beyond conventional shortest-path calculations. Current transportation planning methods often lack the necessary flexibility and personalization to effectively handle such dynamic and complex situations. This paper investigates the potential of Large Language Models (LLMs) to address this deficiency, proposing a novel approach for customized, disruption-aware transportation planning.

\subsection{Limitations of Current Planning Methods}
Traditional route planning primarily utilizes graph search algorithms like Dijkstra's or A* on well-defined transportation networks\cite{fan2010improvement}. While effective for optimization under stable conditions, these methods struggle to dynamically incorporate complex, real-time constraints or qualitative user preferences, such as avoiding specific areas due to perceived danger or discomfort, often expressed colloquially. Research into transportation network resilience\cite{borowska2025assessing, postorino2025new} and emergency routing strategies \cite{kwon2003development, o2024implementing, yu2014routing} provides valuable insights but typically focuses on system-level management or specific responder needs, falling short of the fine-grained personalization required by individual travelers during disruptions. Similarly, while personalized navigation systems aim to incorporate user habits\cite{huang2024personalized}, preferences\cite{ganapathy2021user}, context\cite{yoon2023real, selvaraj2024personalized}, and accessibility needs \cite{yu2012towards}, integrating these features seamlessly with dynamic disruption information and complex avoidance criteria remains an open challenge. Furthermore, effective multimodal planning, integrating public transit with options like bike-sharing \cite{mccoy2018integrating, chavis2016integration, cheng2024bike, zhang2025dynamic, kapica2025synchronization}, faces significant hurdles when needing to adapt dynamically to disruptions and user-specific constraints simultaneously.

\subsection{The Emerging Role of Large Language Models}
Large Language Models (LLMs) such as GPT-4 \cite{achiam2023gpt}, Claude 3 \cite{anthropic2023claude3}, and Gemini \cite{team2023gemini} signify a paradigm shift in artificial intelligence, showcasing remarkable abilities in natural language understanding, reasoning, and complex problem-solving \cite{hadi2023survey, zhao2023survey, wang2024survey}. Their capacity for few-shot learning, in-context reasoning, and processing diverse information formats makes them promising for tasks requiring nuanced understanding and planning \cite{chen2025survey, sui2025stop, aghzal2025survey, li2025planet}. Consequently, LLMs are increasingly explored in various fields, including autonomous driving for scenario interpretation \cite{li2025applications, zhou2023vision, li2024driving}. Within transportation and planning, LLMs are emerging as powerful tools for tasks like developing agents \cite{zhang2025survey, zhao2024semantic, chen2024efficient}, generating personalized routes \cite{marcelyn2025pathgpt}, enhancing pathfinding \cite{meng2025llm, xiao2025llm}, modeling traffic \cite{yan2025large,liu2025llm}, generating mobility patterns \cite{jiawei2024large}, and controlling traffic signals \cite{lai2023llmlight,Masri2025}. These applications highlight the potential of LLMs to grasp spatial concepts \cite{Yang2025}, process complex instructions, and generate contextually relevant mobility outputs.

\begin{figure*}[ht]
    \centering
    \vspace{0.2cm}
    \includegraphics[width=0.7\textwidth]{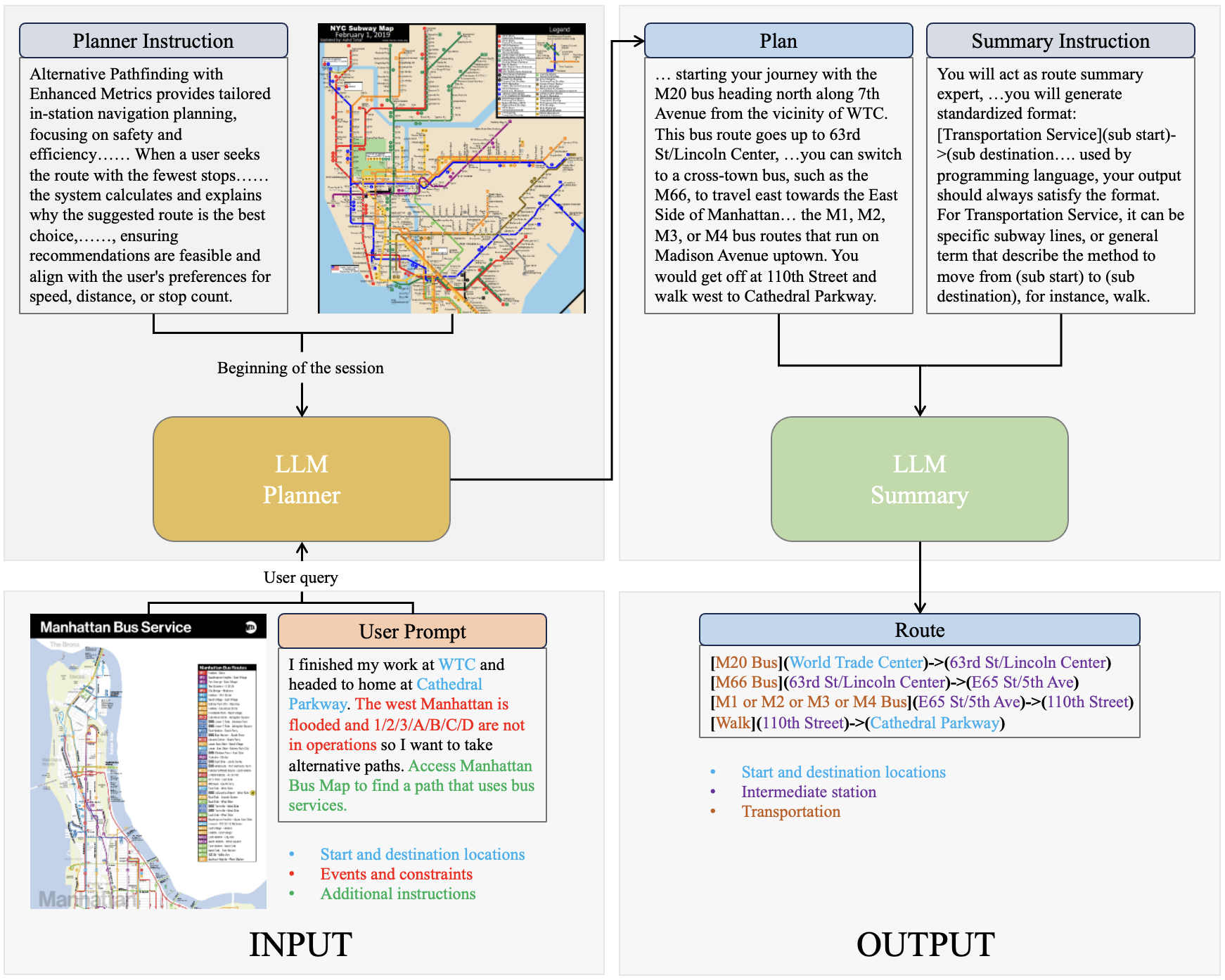} % Adjust path/width as needed
    \caption{System architecture for the TraveLLM prototype, showing the two-stage process from user query to structured plan.}
    \label{fig:outline}
\end{figure*}

Despite these advancements, a critical gap exists in leveraging LLMs specifically for real-time, personalized, multimodal route planning during disruptions. While existing research addresses aspects like personalization, resilience, multimodality, or LLM-based pathfinding individually, the synthesis of these elements remains largely unexplored. Specifically, using an LLM to interpret a natural language query detailing a disruption, personal constraints (e.g., avoiding a windy station exit), and multimodal preferences to generate a feasible path using real-time data is a key challenge. Current systems struggle to fluidly handle the combination of dynamic service changes, arbitrary avoidance criteria, and multimodal integration based on conversational input. To address this gap, we introduce TraveLLM, an approach utilizing the natural language understanding and reasoning capabilities of LLMs. TraveLLM aims to generate customized transportation path recommendations by directly processing user queries, disruption information, and map data, interpreting user intent and context to suggest viable multimodal paths respecting dynamic constraints.

The main contributions of this work are:
\begin{enumerate}
\item A novel LLM-based methodology for personalized transportation planning tailored to handle dynamic disruptions and complex user requirements expressed in natural language.
\item A benchmark suite of test cases based on realistic disruption scenarios (weather, emergencies, service changes) for rigorously evaluating LLM performance in dynamic planning.
\item A comparative analysis of state-of-the-art LLMs, assessing their effectiveness, limitations, and differential capabilities in generating travel recommendations under diverse and challenging conditions.
\end{enumerate}

\section{METHODOLOGY}
This section details the proposed \textbf{TraveLLM} methodology for generating personalized travel path recommendations, particularly designed to handle network disruptions based on multimodal information and user queries.

\subsection{TraveLLM Framework Overview}

TraveLLM generates a structured travel path, $\mathcal{P}_{summary}$, that is feasible, personalized, and disruption-aware. The system takes several inputs (Fig.~\ref{fig:outline}) and employs a two-stage LLM architecture.

The primary inputs informing the process are:
\begin{itemize} 
    \item \textbf{Instructions/Constraints ($\mathcal{I}$):} General guidelines like optimization objectives (e.g., minimize stops). Format: Natural Language (NL).
    \item \textbf{Transportation Services ($\mathcal{S}$):} Descriptions of available network components (e.g., subway/bus maps). Format: Image (IMG) or structured data. Forms part of the knowledge base $\mathcal{KB}$.
    \item \textbf{User Situation/Demands ($\mathcal{U}$):} Query specifics including origin, destination, and preferences. Format: NL.
    \item \textbf{Disruption Information ($\mathcal{D}$):} Details on network issues like service outages or hazards. Format: NL or IMG.
\end{itemize}

Typically, the user query $Q_{user}$ encapsulates $\mathcal{U}$, $\mathcal{D}$, and relevant parts of $\mathcal{I}$. The knowledge base $\mathcal{KB}$ contains $\mathcal{S}$ and global constraints from $\mathcal{I}$. The two LLM stages operate sequentially:

\begin{enumerate}
    \item \textbf{LLM Planner ($\mathcal{L}_{planner}$):} Reasons over the query $Q_{user}$ and knowledge base $\mathcal{KB}$ to generate a detailed natural language plan $\mathcal{P}_{detailed}$.
    \begin{equation}
        \mathcal{P}_{detailed} = \mathcal{L}_{planner}(Q_{user}, \mathcal{KB})
    \end{equation}

    \item \textbf{LLM Summary ($\mathcal{L}_{summary}$):} Parses $\mathcal{P}_{detailed}$ and structures it into a predefined, concise format $F_{std}$.
    \begin{equation}
        \mathcal{P}_{summary} = \mathcal{L}_{summary}(\mathcal{P}_{detailed}, F_{std})
    \end{equation}
\end{enumerate}

We opt for this two-module design ($\mathcal{L}_{planner}$, $\mathcal{L}_{summary}$) using prompt engineering. Our rationale is empirical: enforcing complex planning and rigid formatting ($F_{std}$) simultaneously often degrades plan ($\mathcal{P}_{detailed}$) quality in a single LLM. Decoupling allows each agent to focus effectively. This simple, prompt-based setup works well, avoiding the significant overhead of fine-tuning. We validate this in the Ablation Study.

\subsection{Prompt Engineering Strategy}

We guide $\mathcal{L}_{planner}$ and $\mathcal{L}_{summary}$ using simple, structured prompts.

\textbf{Planner Prompt ($\mathcal{L}_{planner}$):} We instruct $\mathcal{L}_{planner}$ to generate a travel plan using the knowledge base ($\mathcal{KB}$) and the user query ($Q_{user}$ detailing $\mathcal{U}$, $\mathcal{D}$, and $\mathcal{I}$). Key instructions are to prioritize safety (avoiding hazards/disruptions) then efficiency (per $\mathcal{I}$), grounded in $\mathcal{KB}$.

\textbf{Summary Prompt ($\mathcal{L}_{summary}$):} We instruct $\mathcal{L}_{summary}$ to parse the planner's output ($\mathcal{P}_{detailed}$) and generate a structured summary adhering strictly to format $F_{std}$. The prompt forbids extraneous text. It includes examples of transport modes to ensure completeness (e.g., including 'walk'), addressing an observed omission issue.

This straightforward prompt engineering proves effective for controlling the LLM agents.

\section{EXPERIMENTS}

\subsection{Implementation Details}

For the LLM Planner ($\mathcal{L}_{planner}$) component, we evaluate and compare three major large language models known for strong reasoning and multimodal processing capabilities: \textbf{GPT-4} (\texttt{gpt-4-0613}) \cite{achiam2023gpt}, \textbf{Claude 3 Opus} (\texttt{claude-3-opus-20240229}) \cite{anthropic2023claude3}, and \textbf{Gemini Pro 1.0} (\texttt{gemini-1.0-pro}) \cite{team2023gemini}.

For the LLM Summary ($\mathcal{L}_{summary}$) component, which demands reliable instruction-following to generate the structured output format $F_{std}$, we consistently utilize \textbf{GPT-4}.

Our entire TraveLLM approach relies solely on prompt engineering with these pre-trained foundation models. No task-specific training or fine-tuning was performed.

\subsection{Benchmark Scenarios} % Consider renaming for stronger impact

We design a benchmark suite comprising diverse scenarios to rigorously evaluate the capabilities of LLMs in disruption-aware routing. Table~\ref{tab:scenarios} details each scenario. The suite is constructed to probe key aspects, including:
\begin{itemize}
    \item \textbf{Reasoning under Disruptions:} Scenarios involve various disruption types (e.g., extreme weather, emergency events) affecting different subway line geometries (north-south, cross-river, cross-town).
    \item \textbf{Constraint Handling:} Testing the ability to incorporate complex constraints like physical area avoidance (specified textually or visually) and path optimization criteria.
    \item \textbf{Multimodal Information Processing:} Evaluating the integration of supplementary information, such as bus maps or bike-share availability data provided via images.
    \item \textbf{Generalizability:} Assessing performance on a different transit system (Washington D.C. Metro) to test robustness beyond potentially data-rich NYC environments.
\end{itemize}
Each scenario listed in Table~\ref{tab:scenarios} specifies the context and the primary LLM capability under examination.

\begin{table*}[ht]
\centering
\vspace{0.2cm}
\caption{Benchmark Scenario Descriptions and Objectives}
\label{tab:scenarios}
\begin{tabular}{c|p{4.5cm}|p{6.5cm}} % Adjusted column widths
\hline
\hline
Scenario ID & Scenario Description & Test Objective \\
\hline
S1 & North-South Subway Disruption (Extreme Weather) & Test reasoning on general locations during major service outages (West Manhattan flooding, multiple subway lines down). \\
\hline
S2 & Cross-River Subway Route (Location Avoidance) & Test reasoning based on physical network constraints while avoiding a specific, named location (Times Square). \\
\hline
S3 & Cross-Town Subway Disruption (Emergency Event Impact) & Test understanding of how a localized emergency event (attack at Times Square) disrupts a specific planned transfer. \\
\hline
S4 & Cross-Town Subway Route (Emergency Event Non-Impact) & Test reasoning to determine if a nearby emergency event affects the planned route when the specific transfer point is different. \\
\hline
S5 & Subway Route with Visual Avoidance Zone & Test understanding and reasoning based on visual information (marked dangerous zone on a map) requiring route deviation. \\
\hline
S6 & Multimodal Planning (Subway, Bus, Citi Bike) & Test integration of additional transport modes (Citi Bike via image) and handling complex constraints (mode preference, area avoidance for specific mode). \\
\hline
S7 & Subway and Bus Integration (Conditional Map Data) & Test impact on path recommendation when provided with supplementary map data (Manhattan Bus Map) during subway disruption. \\
\hline
S8 & Subway Route with Quantitative Optimization Constraint & Test ability to incorporate quantitative constraints (e.g., minimize stops by preferring express trains) alongside avoidance criteria. \\
\hline
S9 & Non-NYC Subway System (Generalizability Test) & Test generalizability of routing and avoidance reasoning on a different transit system (Washington D.C. Metro). \\
\hline
\hline
\end{tabular}
\end{table*}

\subsection{Evaluation Metrics}\label{sec:metrics}

We evaluate the quality of generated paths 
$\mathcal{P} = (p_0, p_1, ..., p_n)$, where $p_0=O$ (Origin) and $p_n=D$ (Destination), using the following metrics:

\begin{enumerate}
    \item \textbf{Connectivity ($M_{conn}$):} Checks if the path is feasible according to the static transportation network structure provided in the knowledge base $\mathcal{KB}$. Let $Available(p_{i-1}, p_i, m_i)$ be a boolean function indicating if the proposed mode $m_i$ physically connects point $p_{i-1}$ to $p_i$ based on $\mathcal{KB}$.
    \begin{equation}
        M_{conn}(\mathcal{P}) = \bigwedge_{i=1}^{n} Available(p_{i-1}, p_i, m_i) \in \{0, 1\}
    \end{equation}
    A value of $1$ indicates the path is fully connected according to the available services.

    \item \textbf{Constraint Avoidance ($M_{avoid}$):} Determines if the path $\mathcal{P}$ successfully avoids all user-specified avoidance criteria $\mathcal{A}$ (which could be areas, stations, specific lines, etc.).
    \begin{equation}
        M_{avoid}(\mathcal{P}) = \mathbb{I}(\mathcal{P} \cap \mathcal{A} = \emptyset) \in \{0, 1\}
    \end{equation}
    where $\mathbb{I}(\cdot)$ is the indicator function. $M_{avoid}=1$ if all constraints in $\mathcal{A}$ are successfully avoided by the entire path $\mathcal{P}$ (including intermediate points and segments).

    \item \textbf{Normalized Travel Time ($M_{time}$):} Approximates the total travel time $T_{path}(\mathcal{P})$ and normalizes it by the direct walking time $T_{walk}(O, D)$. The time for each path segment $T_i$ (from $p_{i-1}$ to $p_i$ using mode $m_i$) is estimated using Google Maps ($T_{GM}$) queried at a fixed reference time (e.g. May 1, 2024, 1:30 PM EST). If the proposed mode $m_i$ is unavailable, the walking time $T_{walk}(p_{i-1}, p_i)$ is substituted for that segment $T_i$.
    The segment time $T_i$ between $p_{i-1}$ and $p_i$ using mode $m_i$ is:
\begin{equation}
    T_i = \begin{cases} T_{GM}(p_{i-1}, p_i, m_i) & \text{if } Available(p_{i-1}, p_i, m_i) \\ T_{walk}(p_{i-1}, p_i) & \text{otherwise} \end{cases}
\end{equation}
where $T_{GM}$ is Google Maps time queried at a fixed reference. Total path time is:
\begin{equation}
    T_{path}(\mathcal{P}) = \sum_{i=1}^{n} T_i
\end{equation}
The normalized time metric is:
\begin{equation}
    M_{time}(\mathcal{P}) = \frac{T_{path}(\mathcal{P})}{T_{walk}(O, D)}
\end{equation}
    Lower values indicate relatively faster paths compared to walking. If an LLM fails to generate a coherent path from $O$ to $D$, or if the path fails connectivity ($M_{conn}=0$) or avoidance ($M_{avoid}=0$), this metric may be assigned a penalty value (e.g., $\ge 1$) for comparative analysis.

    \item \textbf{Number of Transfers ($M_{transfers}$):} Counts the total number of switches between different transport modes or lines ($m_i \neq m_{i-1}$) along the path $\mathcal{P}$.
    \begin{equation}
        M_{transfers}(\mathcal{P}) = \sum_{i=2}^{n} \mathbb{I}(m_i \neq m_{i-1})
    \end{equation}
    Fewer transfers generally indicate higher convenience.
\end{enumerate}

\section{RESULTS}

We evaluate TraveLLM by comparing planner LLMs and ablating key design choices, using metrics from Section~\ref{sec:metrics}: Connectivity ($M_{conn}$), Avoidance ($M_{avoid}$), Normalized Time ($M_{time}$), and Transfers ($M_{transfers}$).

\subsection{LLM Planner Comparison}

We compare GPT-4, Gemini Pro 1.0, and Claude 3 Opus as the planner ($\mathcal{L}_{planner}$). Table~\ref{tab:model_comparison} shows GPT-4 achieves the best balance, leading significantly in $M_{conn}$ (0.78), $M_{avoid}$ (0.78), and $M_{time}$ (0.51). While Gemini Pro 1.0 yields the fewest $M_{transfers}$ (3.00), its performance on other critical metrics is poor. Claude 3 Opus performs intermediately. This suggests GPT-4 is the most effective planner overall for this task.

% Table 1: Model Comparison
\begin{table}[h] % Consider [ht] for better placement
\centering
\caption{Performance comparison of LLMs as Planner ($\mathcal{L}_{planner}$). Higher is better for $\uparrow$, lower for $\downarrow$.}
\label{tab:model_comparison}
\vspace{0.1cm}
\begin{tabular}{l|ccc}
\hline
\hline
\textbf{Metric} & \textbf{GPT-4} & \textbf{Gemini Pro 1.0} & \textbf{Claude 3 Opus} \\
\hline
$M_{conn}$ $\uparrow$ & \textbf{0.78} & 0.11 & 0.56 \\
\hline
$M_{avoid}$ $\uparrow$ & \textbf{0.78} & 0.22 & 0.67 \\
\hline
$M_{time}$ $\downarrow$ & \textbf{0.51} & 0.92 & 0.64 \\
\hline
$M_{transfers}$ $\downarrow$ & 4.00 & \textbf{3.00} & 4.20 \\
\hline
\hline
\end{tabular}
\end{table}

\subsection{Ablation Studies}\label{sec:ablation}

We perform ablation studies to validate two design choices: using map images ($\mathcal{S}$) in the knowledge base $\mathcal{KB}$, and employing a separate summary agent ($\mathcal{L}_{summary}$).

\subsubsection{Importance of Map Images}

We compare the GPT-4 planner with and without access to map images in $\mathcal{KB}$ (Table~\ref{tab:gpt_map_comparison}). User-provided images (e.g., specifying avoidance zones $\mathcal{A}$) are always included. Results show providing maps significantly improves all metrics: $M_{conn}$ (+0.11), $M_{avoid}$ (+0.34), $M_{time}$ (-0.08), and $M_{transfers}$ (-1.00). This underscores the value of visual map context for planning.

% Table 2: Map Ablation
\begin{table}[h] % Consider [ht]
\centering
\caption{Ablation: Impact of map images ($\mathcal{S}$) for GPT-4 Planner.}
\label{tab:gpt_map_comparison}
\vspace{0.1cm}
\begin{tabular}{l|cc}
\hline
\hline
\textbf{Metric} & \textbf{GPT-4 w/ Map} & \textbf{GPT-4 w/o Map } \\
\hline
$M_{conn}$ $\uparrow$ & \textbf{0.78} & 0.67 \\
\hline
$M_{avoid}$ $\uparrow$ & \textbf{0.78} & 0.44 \\
\hline
$M_{time}$ $\downarrow$ & \textbf{0.51} & 0.59 \\
\hline
$M_{transfers}$ $\downarrow$ & \textbf{4.00} & 5.00 \\
\hline
\hline
\end{tabular}
\end{table}

\subsubsection{Efficacy of Separate Summary Agent}

We compare our two-agent design ($\mathcal{L}_{planner} \to \mathcal{L}_{summary}$) against a single agent ($\mathcal{L}_{planner}$) prompted for direct structured output ($F_{std}$). We introduce Format Violations ($M_{format}$, lower is better) to measure adherence to $F_{std}$.

Table~\ref{tab:summarization_comparison} demonstrates the superiority of the two-agent approach. Using a separate $\mathcal{L}_{summary}$ yields substantially better planning metrics ($M_{conn}$, $M_{avoid}$, $M_{time}$) and drastically reduces format violations ($M_{format}$=0.11 vs. 1.00). The single-agent approach produces lower quality plans and fails completely on format adherence, despite yielding fewer transfers. This validates using a separate agent for reliable, structured output.

The small format violation ($M_{format}$=0.11) with two agents was isolated to Scenario S4, where $\mathcal{L}_{summary}$ added extra text, misinterpreting the planner's valid route in light of broad instructions. While indicating a potential interaction issue, the overall benefits of the two-agent design in plan quality and format reliability are clear.

% Table 3: Summarizer Ablation
\begin{table}[h] % Consider [ht]
\centering
\caption{Ablation: Impact of using a separate LLM Summary agent ($\mathcal{L}_{summary}$).}
\label{tab:summarization_comparison}
\vspace{0.1cm}
\begin{tabular}{l|cc}
\hline
\hline
\textbf{Metric} & \multicolumn{2}{c}{\textbf{Separate $\mathcal{L}_{summary}$?}} \\
\cline{2-3}
& \textbf{Yes (Two Agents)} & \textbf{No (Single Agent)} \\
\hline
$M_{conn}$ $\uparrow$ & \textbf{0.78} & 0.44 \\
\hline
$M_{avoid}$ $\uparrow$ & \textbf{0.78} & 0.22 \\
\hline
$M_{time}$ $\downarrow$ & \textbf{0.51} & 0.65 \\
\hline
$M_{transfers}$ $\downarrow$ & 4.00 & \textbf{2.80} \\
\hline
$M_{format}$ $\downarrow$ & \textbf{0.11} & 1.00 \\
\hline
\hline
\end{tabular}
\end{table}

\section{DISCUSSION}

Our results demonstrate the promise of LLMs for disruption-aware routing via TraveLLM, while also highlighting current limitations.

% Add this paragraph to the Discussion section (or integrate it logically)
% Keep Figs for S5 and S6 as they are referenced below.
\begin{figure}[h] % Suggest [ht] for placement flexibility
    \centering
    \includegraphics[width=0.6\linewidth]{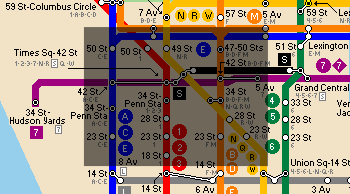} % Adjusted width
    \caption{Visual avoidance constraint for Scenario S5: avoid the black rectangle.}
    \label{subfig:nyc_subway_exp1}
\end{figure}

\begin{figure}[h] % Suggest [ht]
    \centering
    \includegraphics[width=0.35\linewidth]{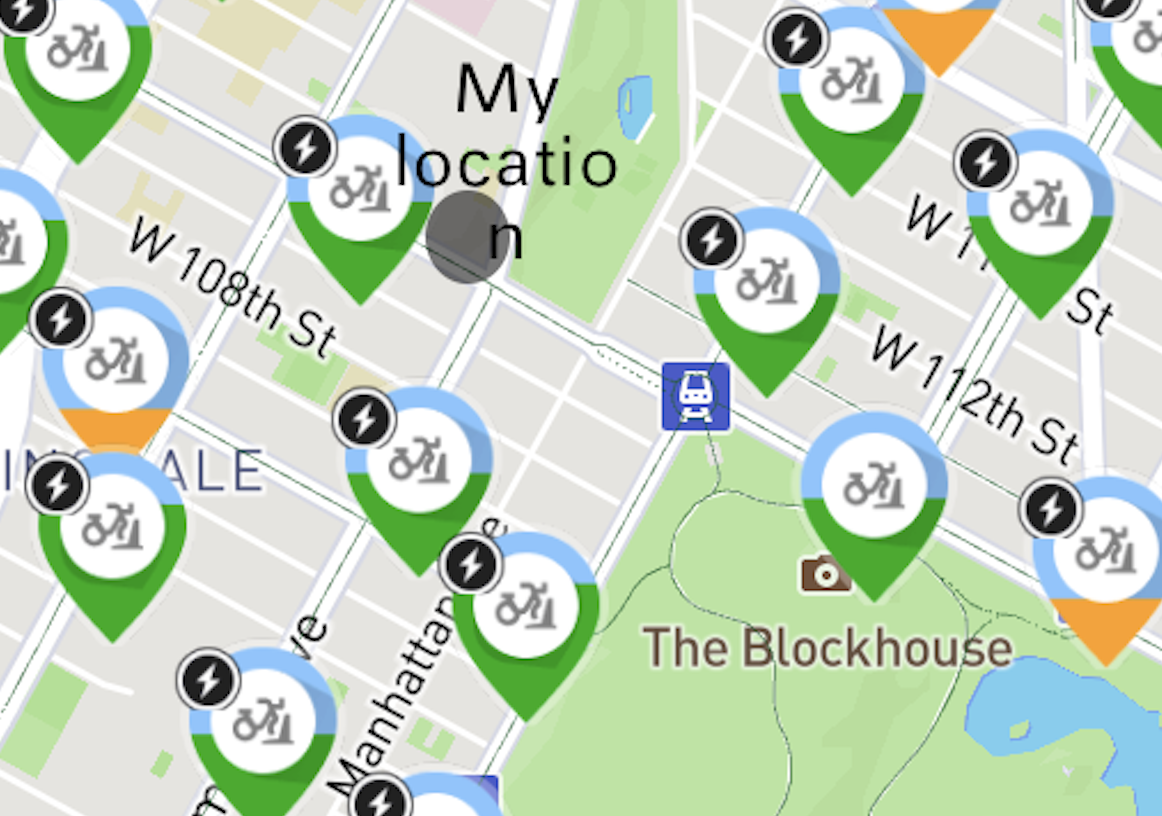} \hspace{0.1cm} % Added space
    \includegraphics[width=0.4\linewidth]{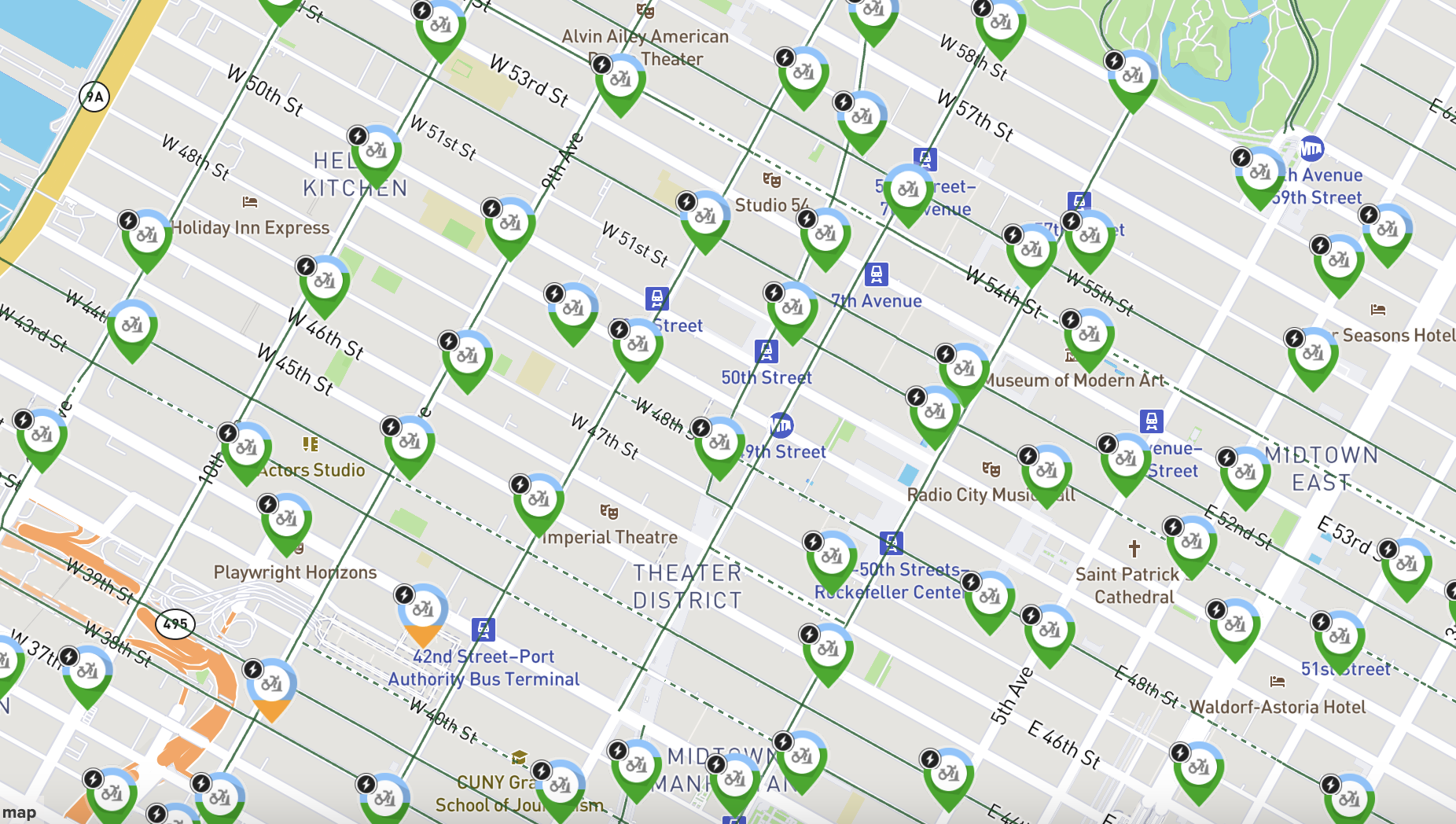} % Adjusted widths
    \caption{Visual input for multimodal Scenario S6: user location (circle) and Citi Bike availability (bubbles).}
    \label{subfig:bike}
\end{figure}
% Removed fig:example figure environment and its label.

\textbf{Limitations.} Current models exhibit limitations in complex reasoning and ensuring basic path feasibility.
First, integrating visual constraints with pathfinding remains difficult. In Scenario S5 (Fig.~\ref{subfig:nyc_subway_exp1}), despite the general benefit of map images shown in our ablation (Table~\ref{tab:gpt_map_comparison}), no model produced a route that simultaneously satisfied the visual avoidance constraint ($M_{avoid}=1$) and maintained path connectivity ($M_{conn}=1$). This suggests challenges in fine-grained visual-spatial reasoning applied to planning.
Second, ensuring basic connectivity ($M_{conn}$) can fail even when simpler constraints are met. For example, in Scenario S2 (avoid Times Square), analysis suggests Claude 3 and Gemini Pro 1.0 generated disconnected paths (violating $M_{conn}$), whereas GPT-4 succeeded, despite all potentially satisfying the primary avoidance constraint.

\textbf{Strengths and Variability.} We observe variability in LLM capabilities, particularly for multimodal tasks. Scenario S6 required planning with transit and bike-sharing, using visual data for bike availability (Fig.~\ref{subfig:bike}) and respecting biking exclusion zones. Claude 3 Opus uniquely handled this complexity well: it identified specific bike stations, interpreted availability cues from the image, planned the rent/return logistics, and generated a compliant multimodal path. In contrast, GPT-4 and Gemini Pro 1.0 failed to effectively incorporate the bike-share details. This highlights differing abilities among models to handle less common transport modes or interpret specific visual information patterns.

\textbf{Practical Considerations.} The practical impact of some errors depends on integration context. Our metrics, like $M_{conn}$ and $M_{transfers}$, require detailed paths. If TraveLLM's output is used to provide high-level guidance (mode, key locations) to another navigation API (e.g., Google Directions), minor errors in specific train line choices between valid points might be implicitly corrected by that API. However, fundamental errors such as proposing disconnected segments or violating critical avoidance constraints ($M_{avoid}=0$) remain significant failures. The validated reliability of our two-stage architecture in producing well-formatted output ($M_{format}=0.11$, Table~\ref{tab:summarization_comparison}) is crucial for enabling such practical integration.

\begin{figure*}[htbp]
    \centering
    % Assume multi_routes.jpg is in the correct path relative to your .tex file
    \includegraphics[width=\textwidth]{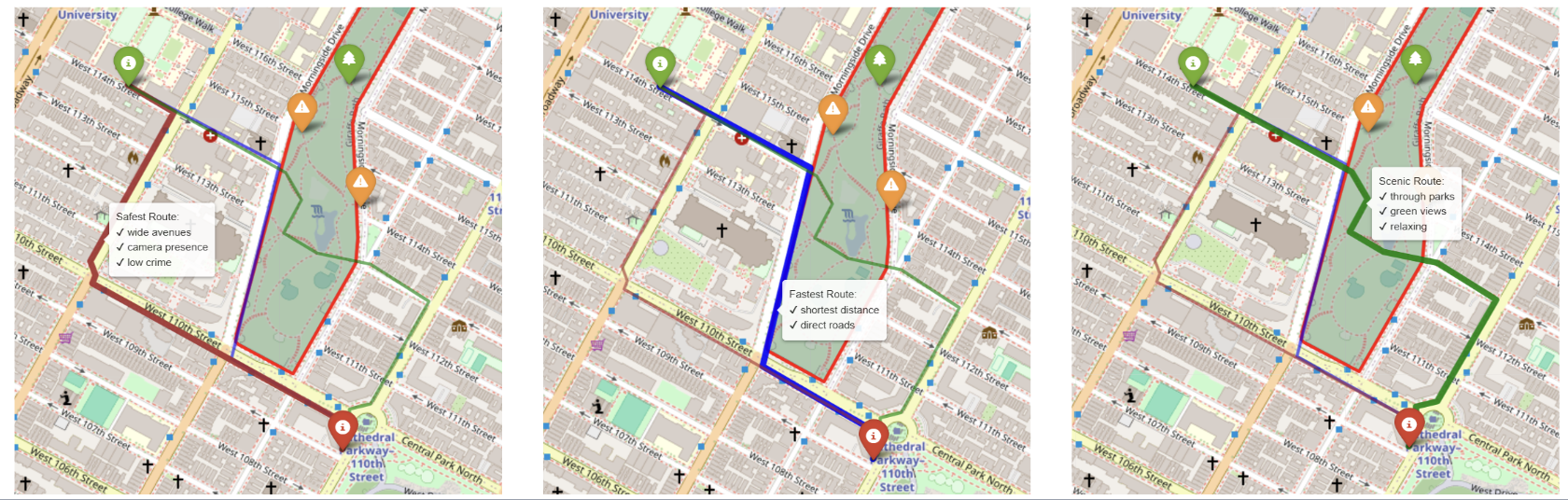}
    \caption{Example of diverse route generation by TraveLLM. For a trip from Columbia University to 110th St station, distinct routes optimizing for safety (left, red), efficiency (middle, blue), and scenery (right, green) are generated based on qualitative criteria interpreted by the LLM Planner ($\mathcal{L}_{planner}$). Route justifications are derived from planner output.}
    \label{fig:multi_routes}
\end{figure*}

\textbf{Generating Diverse, Qualitative Routes.} Beyond handling disruptions, our framework allows LLMs to generate routes based on nuanced, qualitative criteria. Figure~\ref{fig:multi_routes} demonstrates this for a walk from Columbia University to the 110th St subway station. By simply modifying the prompt to prioritize different aspects, TraveLLM produces distinct paths optimized for user preferences like safety, efficiency, or scenery. The 'safest' route (red) utilizes main avenues, the 'fastest' (blue) takes the most direct path, and the 'scenic' (green) route traverses the nearby park. This showcases the ability of the LLM Planner ($\mathcal{L}_{planner}$) to interpret abstract goals, ground them in relevant environmental features (e.g., park attributes vs. avenue characteristics), and generate corresponding path geometries. The textual descriptions justifying each route choice (e.g., "Wide avenues", "Through park"), originating from $\mathcal{L}_{planner}$'s reasoning, can be extracted and formatted by $\mathcal{L}_{summary}$, providing users with explainable and personalized navigation options that go beyond simple time or distance optimization.

\section{CONCLUSION}

In this paper, we presented \textbf{TraveLLM}, an approach leveraging Large Language Models for personalized, disruption-aware public transit routing. Our experiments demonstrated that current LLMs, particularly GPT-4, can effectively generate viable navigation plans under complex scenarios involving disruptions, multimodal data (maps, text), and user-specific constraints. Through ablation studies, we verified the benefits of incorporating visual map information and employing a two-stage planner-summarizer architecture for enhancing plan quality and ensuring reliable, structured output.

While we observed limitations, notably in handling complex visual constraints (e.g., Scenario S5) and integrating less common transportation modes effectively (e.g., Scenario S6), our findings establish the promise of using LLMs for more flexible and adaptive navigation systems. This work provides a foundation for future research focused on improving visual reasoning, ensuring factual grounding, and refining the integration of LLM-based planners into practical navigation applications.

%%%%%%%%%%%%%%%%%%%%%%%%%%%%%%%%%%%%%%%%%%%%%%%%%%%%%%%%%%%%%%%%%%%%%%%%%%%%%%%%

%%%%%%%%%%%%%%%%%%%%%%%%%%%%%%%%%%%%%%%%%%%%%%%%%%%%%%%%%%%%%%%%%%%%%%%%%%%%%%%%

%%%%%%%%%%%%%%%%%%%%%%%%%%%%%%%%%%%%%%%%%%%%%%%%%%%%%%%%%%%%%%%%%%%%%%%%%%%%%%%%

%%%%%%%%%%%%%%%%%%%%%%%%%%%%%%%%%%%%%%%%%%%%%%%%%%%%%%%%%%%%%%%%%%%%%%%%%%%%%%%%

\bibliography{itsc24}

\addtolength{\textheight}{-12cm}   % This command serves to balance the column lengths

\end{document}